%% file: neurips_2025.tex
\documentclass{article}

\usepackage[preprint]{neurips_2025}
\usepackage{enumitem}
\usepackage{booktabs} 
\usepackage{multicol}
\usepackage{graphicx}
\usepackage{subcaption}
\usepackage{amsmath}
\usepackage{amssymb} 
\usepackage{hyperref}
\usepackage{wrapfig}
\usepackage{natbib}
\usepackage{makecell} 
\usepackage{soul, color, xcolor}       
\sethlcolor{green}
\usepackage[ruled,linesnumbered]{algorithm2e}

\usepackage[utf8]{inputenc} 
\usepackage[T1]{fontenc}    
\usepackage{hyperref}       
\usepackage{url}            
\usepackage{booktabs}       
\usepackage{amsfonts}       
\usepackage{nicefrac}       
\usepackage{microtype}      
\usepackage{xcolor}         
\usepackage{colortbl}

\title{KVShare: An LLM Service System with Efficient and Effective Multi-Tenant KV Cache Reuse}

\author{
Huan Yang \\
Central South University\\
\texttt{yanghuan9812@csu.edu.cn} \\
\And
Renji Zhang \\
Central South University\\
\texttt{224707001@csu.edu.cn} \\
\And
Mingzhe Huang \\
Tsinghua University \\
\texttt{huangmz21@mails.tsinghua.edu.cn} \\
\And
Weijun Wang \\
Tsinghua University \\
\texttt{wangweijun@air.tsinghua.edu.cn} \\
\And 
Yin Tang\\
Central South University \\ 
\texttt{yintag@gmail.com} \\
\And
Yuanchun Li \\
Tsinghua University \\ 
\texttt{liyuanchun@air.tsinghua.edu.cn} \\
\And
Yunxin Liu  \\
Tsinghua University \\ 
\texttt{liuyunxin@air.tsinghua.edu.cn} \\
\And
Deyu Zhang\\
Central South University \\
\texttt{zdy876@csu.edu.cn} \\
}

\begin{document}
\let\thanks\relax
\begin{center}

\maketitle

\end{center}

\begin{abstract}
Recent advances in long-text understanding have pushed the context length of large language models (LLMs) up to one million tokens.  It boosts LLMs' accuracy and reasoning capacity but causes exorbitant computational costs and unsatisfactory Time to First Token (TTFT). KV cache reuse, which reuses the exact same KV cache of prefixes and templates or shares similar ones but with extra selective recomputation, offers a promising way to tackle this issue. However, prior studies overlook the cross-request KV reuse and the attention deviations introduced by new tokens during the decoding stage. In this paper, we present a KV cache management module that shares the KV cache across requests under multi-tenant scenarios without sacrificing model accuracy. Our system, KVShare, enables accurate and efficient LLM serving by  1) a Dual-Stage High Deviation algorithm (DHD) that conditionally selects a small portion of KV cache to be recomputed during both prefill and decode phases, and 2) a cache-aware scheduler that prioritizes requests based on their KV cache hit rates and orchestrates continuous batching to achieve enhanced system efficiency and faster TTFT. Multi-task experiments conducted on models such as Qwen2.5-7B,Llama3.1-8B and Yi1.5-9B demonstrate that KVShare reduces TTFT by up to 9.39$\times$ and increases 1.2$\times$ of the throughput compared to the full KV recompute. Moreover, KVShare achieves 20.38\% boost in terms of accuracy compared to SOTA methods. 

\end{abstract}

\input{section/introduction}
\input{section/background}
\input{section/problem}

\input{section/design}

\input{section/system}
\input{section/eval}
\input{section/conclusion}

\bibliographystyle{unsrtnat}
\bibliography{neurips_2025}

\input{section/appendix}
\input{section/checklist}

\end{document}

%% file: section/introduction.tex
\section{Introduction}

Large language models (LLMs) \citep{qwq32b,glm2024chatglm} have made significant progress in context-handling capabilities, with the latest models able to process sequences of up to a 1-million context window.
Such a long context enhances their reasoning ability but causes exorbitant computational overhead, which burdens service providers and significantly slows users' Time to First Token (TTFT). 
Existing context-caching techniques \citep{bang2023gptcache, gim2024prompt, zheng2025sglang} reuse KV cache via common prefixes or customized templates, reducing GPU workload and TTFT. 
However, they fall short in multi-tenant scenarios that require cross-request KV reuse.

The restriction on identical prefixes in the prior KV cache reuse mechanism misses a large opportunity for KV cache reuse.
Since the auto-regressive nature of LLMs, KV cache is inherently dependent on the preceding prefix tokens. 
Given two requests including an identical token chunk, their KV caches may still differ from each other due to the distinct prefixes. 
Under this, naively reusing the KV cache of this token chunk from one request to another, without considering their different prefixes, will cause a deviation in the attention matrix (so-called \emph{attention deviation}) compared to the original one without KV cache reuse.


Existing fixed-length token chunking hinders the KV cache reuse.
Recent works such as CacheBlend \citep{yao2024cacheblend} and EPIC \citep{hu2024epic}, mitigating attention deviation by selecting and recomputing the KV cache of a bunch of critical tokens in each chunk, has made noticeable progress.
However, their fixed chunk size can only deliver suboptimal cache reuse in multi-tenant LLM serving. 
This is because the reusable token chunks are often non-contiguous and of different sizes across user requests.
Therefore, reusing the KV cache with a dynamic chunk size is critical.

\begin{wrapfigure}{br}{0.45\textwidth} 
 \vspace{-1em}
 \centering
    \includegraphics[width=0.9\linewidth]{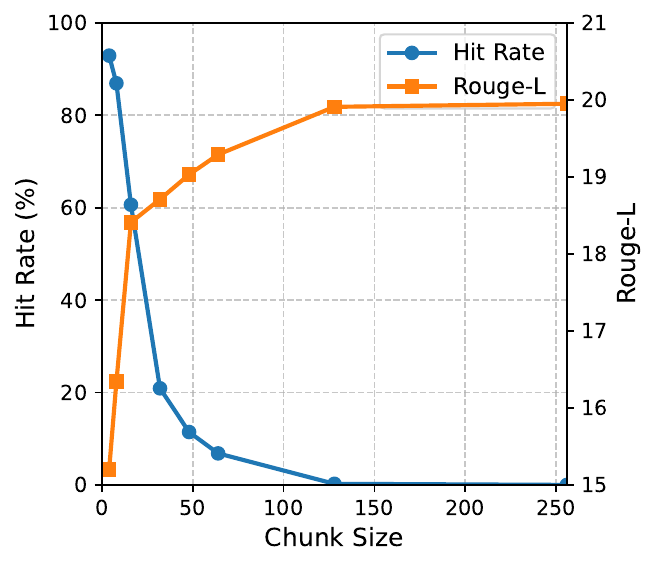}
    \caption{The relation between chunk size and accuracy \& hit rate. The results are profiled with Llama3.1-8B in SAMSum dataset.}
    \label{fig:the_accuracy_and_hit_rate_trade_off}
    \vspace{-1em}
\end{wrapfigure}

To this end, our goal is to develop an adaptive-length token chunking strategy that can reuse KV cache as much as possible while minimizing the attention deviation across all user requests, which poses two inherent challenges:
(1) Changing relation between chunk size and token selection algorithm accuracy.
As shown in Figure \ref{fig:the_accuracy_and_hit_rate_trade_off}, smaller chunk sizes yield more likely cache reuse (a.k.a. cache hit rate) while suffering from a low algorithm accuracy, and vice versa. There is no one-size-fits-all chunk size for cache reuse algorithm. The algorithm must dynamically adapt to varying chunk sizes across different requests and establish an optimized Pareto frontier for effective performance balancing (more analysis in \S\ref{section:4.1} \& \S\ref{section:4.2}).
(2) Inefficient scheduling under significantly different cache hit rates across multi-tenant requests.
A high hit rate results in more KV cache reuse and lower computing costs, while a low hit rate requires more computing costs.
Unfortunately, the cache hit rates of user requests vary in multi-tenant scenarios. 
Hit-rate-agnostic scheduling prevents high-hit-rate requests from early output, leading to a high TTFT (more analysis in \S\ref{sec:scheduling}).

This paper presents KVShare, a flexible and efficient multi-tenant KV cache sharing module for LLM serving systems, as shown in Figure \ref{fig:kvshare_system}. KVShare first employs an adaptive-length token reuse strategy,as shown in Appendix \ref{appendix:E} within the KV Retriever to improve cache hit rates. Building on this, a cache-aware scheduler reduces TTFT in multi-tenant scenarios. Finally, the Dual Stage High Deviation (DHD) selector restores attention fidelity in both prefill and decoding stages, enhancing token recomputation accuracy and maintaining generation quality.



To evaluate the efficiency and effectiveness of KVShare, we conduct extensive experiments on Qwen2.5-7B \citep{qwen2.5} , Llama3.1-8B \citep{llama3modelcard} and Yi1.5-9B \citep{young2024yi} across diverse types of tasks. We select Full Recompute (FR) and Naive as our baseline. The FR baseline completely recomputes all tokens, while the Naive baseline completely reuses the KV cache of all tokens. Additionally, we compare KVShare with state-of-the-art (SOTA) methods \citep{yao2024cacheblend,hu2024epic}. The experimental results demonstrate that KVShare reduces TTFT by up to 9.39$\times$ and increases the throughput by 1.2$\times$, compared with full KV recompute. Moreover, KVShare achieves 20.38\% boost in terms of accuracy compared to SOTA methods.

In summary, our main contributions are listed below:



\begin{enumerate} 
    \item We propose KVShare, a fine-grained multi-tenant KV cache sharing framework that reduces TTFT by up to 9.39$\times$ and boosts 1.2$\times$ of the throughput compared to the full KV recompute.
    \item We propose a cache-aware scheduler that reduces TTFT by up to 1.5$\times$ with negligible additional overhead. 
    \item We provide a detailed analysis of the attention mechanism, and propose our Dual Stage High Deviation (DHD) algorithm, which achieves an average of 20.38\% accuracy improvement over the state-of-the-art method CacheBlend.
\end{enumerate}

\begin{figure*}[htbp]
  \vspace{-1em}
  \centering
  \includegraphics[width=0.7\linewidth]{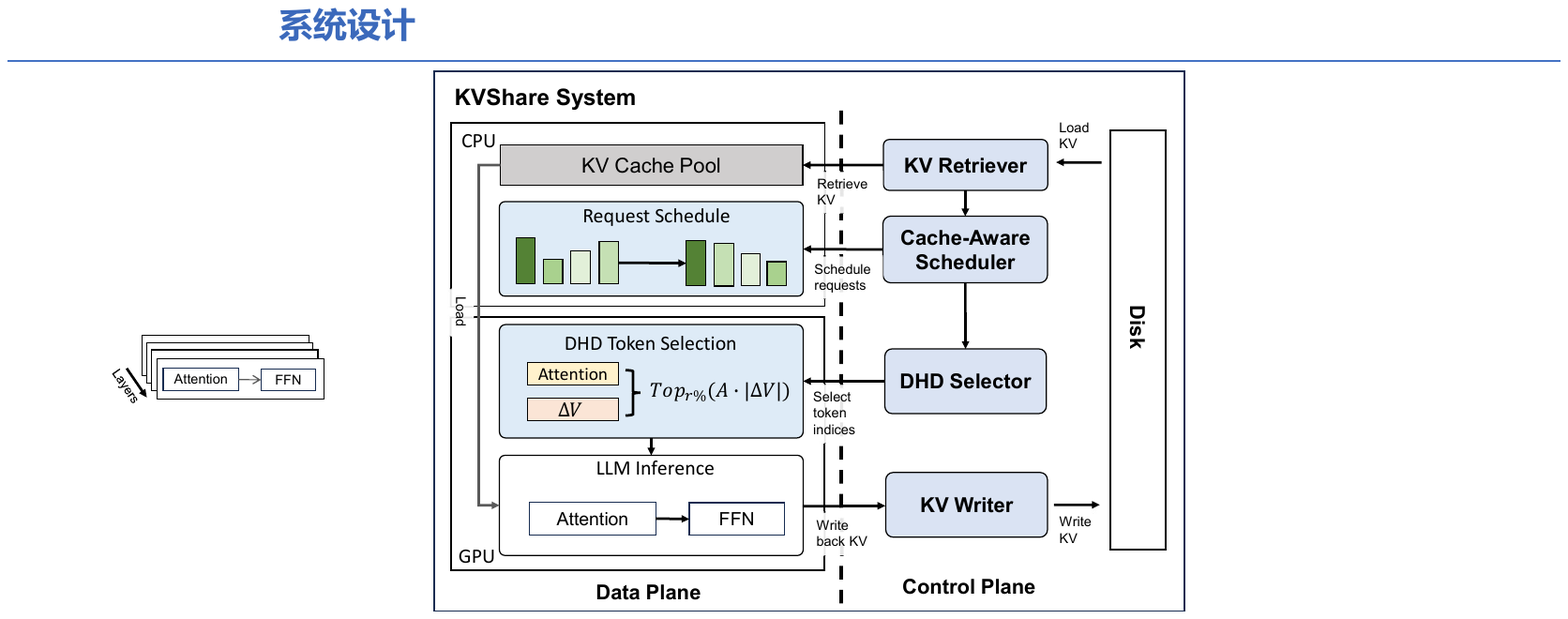}
  \caption{KVShare system for low-latency context retrieval and adaptive token re-computation. }
  \label{fig:kvshare_system}
  \vspace{-1em}
\end{figure*}

%% file: section/background.tex
\section{Related Work}

\textbf{Prefix Caching} 
\citep{qin2024mooncake,zheng2025sglang} shares the KV cache corresponding to the same prefix text in multi-user requests. 
In a multi-tenant environment, many users often use the same system prompts or deal with a large amount of text with the same prefix (such as web content). 
By reusing these KV caches through Prefix Cache, it can significantly reduce the computational burden of the attention mechanism module in the large language model. 
However, this approach is limited to scenarios with exact prefix matches and struggles to handle requests with structural variations.

\textbf{Context Caching} \citep{gim2024prompt, yao2024cacheblend, hu2024epic, kwon2023efficient, zheng2025sglang} performs chunking on the prompt, achieves the reuse of KV cache for non-prefix text through exact semantic matching between chunks, and recalculates partial tokens to restore the accuracy of the large language model (LLM). Specifically,  Epic \citep{hu2024epic} restores accuracy by minimizing the recomputation overhead through static attention sparsity, while Cacheblend \citep{yao2024cacheblend} mitigates attention loss by selectively recomputing a subset of tokens with high KV cache deviation.
However, such methods rely on accurate estimation of deviation distributions and often suffer from low cache hit rates in multi-user scenarios.


%% file: section/problem.tex
\section{Problem}
\label{section:3.1}

\begin{wrapfigure}{br}{0.45\textwidth} 
 \vspace{-1em}
 \centering
    \includegraphics[width=0.9\linewidth]{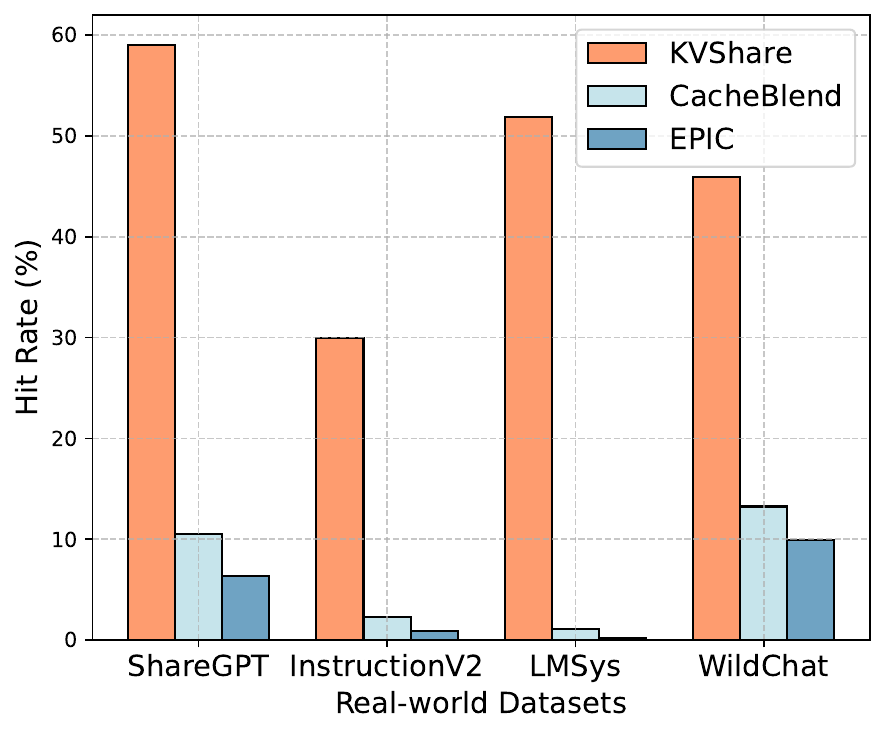}
    \caption{Comparison of hit rates for KVShare, CacheBlend, and EPIC methods across four chat datasets: ShareGPT \citep{ShareGPT-Chinese-English-90k}, InstructionV2 \citep{instructionwild}, LMSys\citep{instructionwild}, and WildChat \citep{zhao2024wildchat}.}
    \label{fig:hit_rate_comparison}
    \vspace{-2em}
\end{wrapfigure} 

As illustrated in Figure \ref{fig:hit_rate_comparison}, recent algorithms such as CacheBlend \citep{yao2024cacheblend} and EPIC \citep{hu2024epic}  suffer from low cache hit rates in multi-tenant scenarios due to their reliance on fixed token chunk size and strict token block matching requirements. 

Consequently, their ability to leverage KV cache reuse—a key mechanism for reducing TTFT—is significantly diminished in multi-tenant scenarios. Moreover, these methods fail to account for the performance imbalance introduced by heterogeneous cache hit rates during batch execution, where the latency advantages of high-hit-rate requests are often masked by the overhead incurred from low-hit-rate requests, ultimately constraining the overall system efficiency. To overcome the limitations of previous approaches, we focus not just on improving cache hit rates, but also on addressing two critical challenges: (1) maintaining accuracy in cross-request KV cache reuse, and (2) developing effective task scheduling strategies that account for varying cache hit rates across requests.


\textbf{Changing relation between chunk size and token selection algorithm accuracy}
Cross-request KV cache reuse reduces redundant computation via sharing identical token fragment KV caches. To mitigate attention degradation from mismatched reused prefixes, selective token recomputation restores accuracy, aligning with KV deviation goals \citep{zhang2023h2o,li2024snapkv}. 
CacheBlend \citep{yao2024cacheblend} prioritizes tokens by KV deviation magnitude but overlooks that largest deviations do not always correlate with highest attention, limiting effective token identification. EPIC \citep{hu2024epic} builds on this by observing sink tokens \citep{xiao2023efficient} (often at document beginnings) exhibit high attention and large KV deviations, though it cannot distinguish intra-block token importance or handle variable reused chunk lengths in practice.

The existing solutions fall short in providing a comprehensive analysis of how attention deviation arises from K and V biases across Transformer layers. This limits their ability to efficiently and accurately guide token selection for recomputation.



\textbf{Inefficient scheduling under significantly different cache hit rates across multi-tenant requests} 

In multi-tenant scenarios, continuous batching techniques are commonly used to merge prefill requests and improve GPU utilization and input throughput. However, traditional scheduling strategies group prefill requests with high and low KV hit rates into the same batch \citep{kwon2023efficient} , failing to account for differences in KV hit rates among requests. 
The processing time advantage of high hit rate requests is diluted by the prolonged execution of low hit rate requests, increasing the average TTFT and reducing throughput.Such inefficient task scheduling hampers resource allocation, reducing utilization under concurrent multi-requests and significantly degrading overall system performance.

%% file: section/design.tex
\section{Methodology}

To address the performance degradation in LLMs caused by KV cache reuse, we propose a two-stage optimization strategy. First, we incorporate attention weights into the formulation of attention output bias, enabling more effective restoration of attention fidelity. 
Second, we design a dual-stage high KV deviation token recomputation mechanism that selectively recomputes critical tokens during the prefill and decode stages, thereby mitigating accuracy loss with minimal additional computational overhead.
Furthermore, to resolve the performance masking effect caused by heterogeneous cache hit rates among concurrent requests, we design a simple yet effective scheduling strategy. By prioritizing requests based on their cache hit rates and applying continuous batching accordingly, our method maximizes the benefits of KV cache reuse and enhances overall system performance. 


\begin{algorithm}[htbp]
\caption{Dual Stage High Deviation Token Selection}
\label{alg:deviation_aware}
\KwIn{Query matrix $Q$, Key matrix $K$, Value matrix $V$; \\
\quad\ \ Deviation matrix $\Delta V$; \\
\quad\ \ Recompute ratio $r\%$}
\KwOut{Indices of tokens selected for recomputation}

Compute attention weights: 
\quad $A \gets \mathrm{Softmax}\left(\frac{Q K^\top}{\sqrt{d_k}}\right)$\;

Compute per-token cumulative attention: 
\quad $\alpha \gets \mathrm{ColSum}(A)$\; 

Compute deviation impact scores: 
\quad $\text{Score} \gets \alpha \cdot \left\| \Delta V \right\|_1$\;

Select top-$r\%$ tokens with highest $\text{Score}[i]$ as recompute candidates\;

\Return{Selected token indices}
\end{algorithm}

\subsection{Critical Tokens Depend on the Attention}
\label{section:4.1}

In the Transformer architecture, reusing the Key-Value Cache of certain tokens effectively introduces increments $\Delta K$ and $\Delta V$ to the original Key ($K$), and Value ($V$) matrices, as shown in Equation \eqref{eq:attention_formula}.

\begin{equation}
\begin{aligned}
H = \mathrm{Softmax}\left(\frac{QK^T}{\sqrt{d_k}}\right)V, \quad H' =  \mathrm{Softmax}\left(\frac{Q(K+\Delta K)^T}{\sqrt{d_k}}\right)(V+\Delta V)
\end{aligned}
\label{eq:attention_formula}
\end{equation}

This operation transforms the original correct attention result $H$ into $H'$, thereby inducing an  deviation $\Delta H$. To investigate the intrinsic relationships among $\Delta K$, $\Delta V$, and $\Delta H$, we perform a first-order Taylor expansion on $\Delta H$ and derive Equation \eqref{eq:attention_differentiation}.


\begin{figure*}[htbp]
    \centering
    \begin{subfigure}[t]{0.45\textwidth} 
        \includegraphics[width=0.9\linewidth]{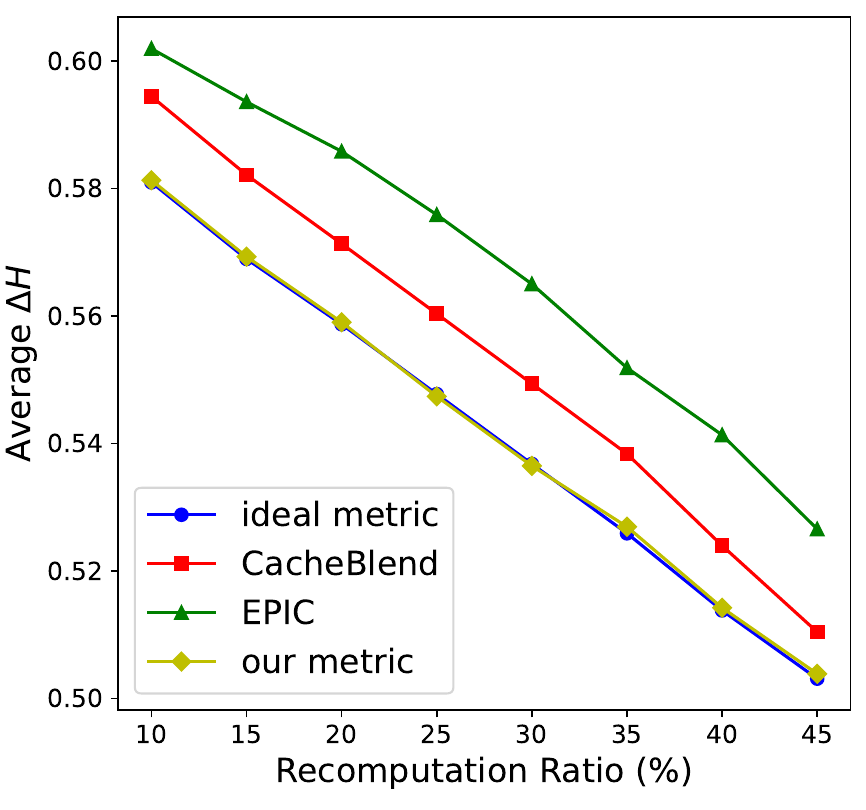}
        \caption{The impact of different token recomputation strategies on $\Delta H$.}
        \label{fig:recompute_strategy_comparison}
    \end{subfigure}
    \hfill
    \begin{subfigure}[t]{0.45\textwidth} 
        \includegraphics[width=0.9\linewidth]{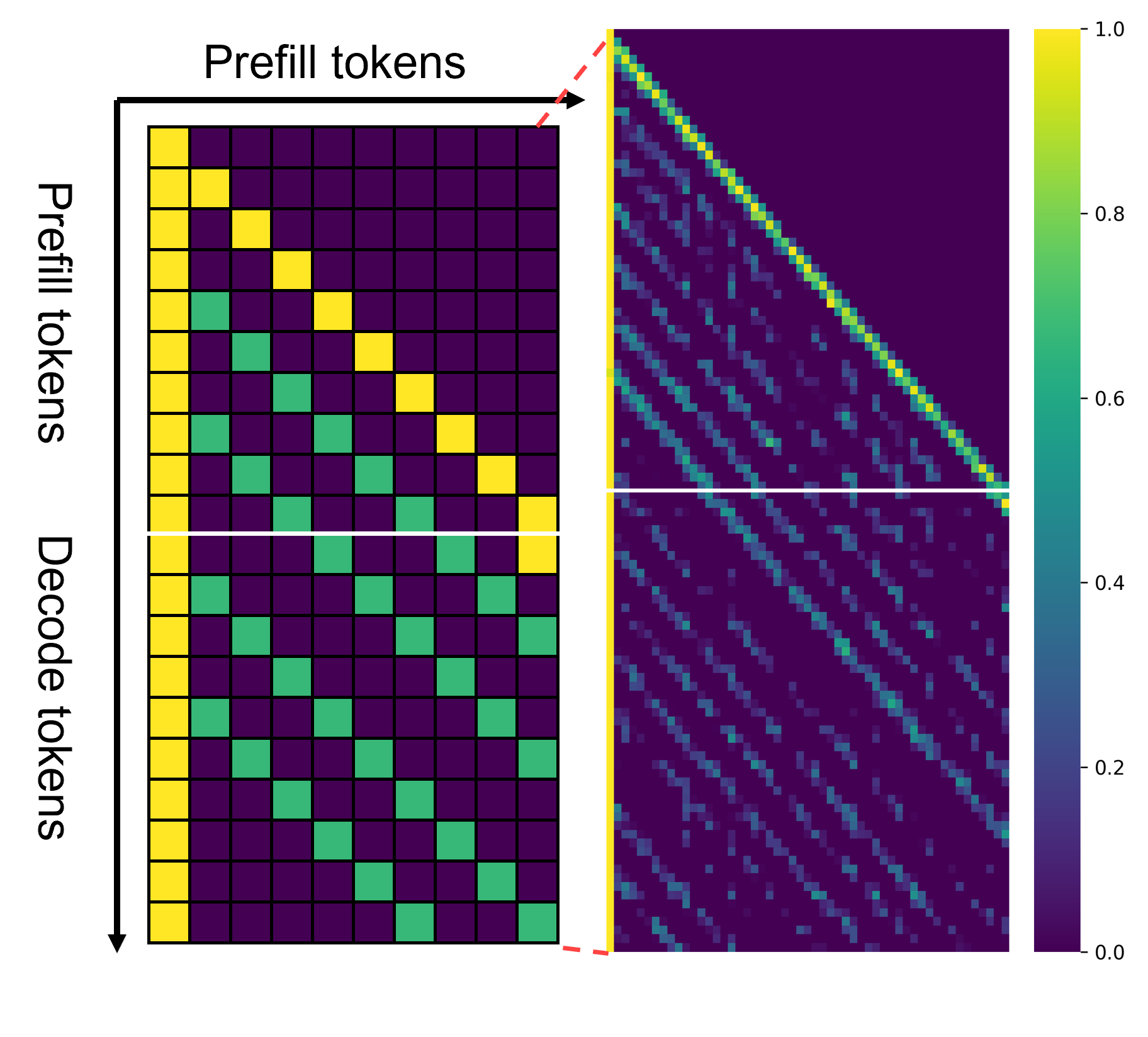}
        \caption{Attention shift in the decode stage.}
        \label{fig:attention_shift}
    \end{subfigure}
    \caption{In Figure\ref{fig:recompute_strategy_comparison}, we use the Llama3.1-8B model on the GSM8K dataset to test the impact of different token selection strategies on the error of $\Delta H$. The results show that the ideal index method and our proposed method have similar effects, both of which can minimize the error of $\Delta H$ to the greatest extent. In Figure\ref{fig:attention_shift}, it is evident that there is a significant shift in the attention during the decode phase compared to the cross - attention in the prefill phase. This indicates that some tokens with relatively low attention scores in the prefill phase receive higher attention during the decode phase. We visualized this using the attention data from 12 heads in Layer 25 of Llama 3.1.}
    \label{fig:exec_batch_hit}
\end{figure*}

Equation \eqref{eq:attention_differentiation} reveals that the value of $\Delta H$ is determined not only by $\Delta K$ and $\Delta V$, but also by the partial derivatives $\frac{\partial H}{\partial K}$ and $\frac{\partial H}{\partial V}$. 
Specifically,The attention matrix itself serves as the Jacobian, i.e., \( \frac{\partial H}{\partial V} = A \), modulating how deviations in \( V \) propagate to \( H \).

\begin{equation}
\begin{aligned}
\Delta H &= H' - H \approx  \frac{\partial H}{\partial K} \Delta K + \frac{\partial H}{\partial V} \Delta V \\
\end{aligned}
\label{eq:attention_differentiation}
\end{equation}

As shown in Figure~\ref{fig:recompute_strategy_comparison}, we quantify the impact of different token selection strategies by measuring the attention output deviation induced under the same recomputation budget during inference. Our results show that the strategy guided by the attention Jacobian—specifically using the first-order approximation $\frac{\partial H}{\partial K} \Delta K + \frac{\partial H}{\partial V} \Delta V$—achieves the lowest attention error. In contrast, position-based heuristics like EPIC and deviation-magnitude-based approaches like Blend perform worse, as they fail to fully account for how the key and value deviations propagate through the attention mechanism to affect the output.

Furthermore, we analyze the token sets ranked by their key- and value-directional impacts and find a substantial overlap of 80–90\% between them,as shown in Appendix \ref{appendix:D.2}, indicating strong consistency between the two directions. To reduce computational overhead, we simplify the selection process by considering only the value-directional component, i.e., $\frac{\partial H}{\partial V} \Delta V$, and empirically find that this approximation achieves nearly identical accuracy to the full Jacobian-based method. Motivated by this insight, we develop a new algorithm \ref{alg:deviation_aware} for selecting tokens to recompute.



As shown in Equation \eqref{eq:dhd_formula},  we not only consider the inherent bias in the value vectors but also the relative importance of each token as determined by its attention weight. This aligns well with our mathematical intuition and offers a computationally efficient yet effective proxy for identifying tokens that most critically impact attention accuracy.

\begin{equation}
\begin{aligned}
\text{Score}_i = \left( \sum_{j=1}^N \mathrm{Softmax}\left(\frac{Q_j K_i^\top}{\sqrt{d_k}}\right) \right) \cdot \left\| \Delta V_i \right\|_1
\end{aligned}
\label{eq:dhd_formula}
\end{equation}

\subsection{Reassessing Token Importance During the Decoding Phase}
\label{section:4.2}


In decoding, the reused KV cache from earlier tokens influence the attention computation of each newly generated token, and each new token in turn contributes to future attention computations. As a result, any bias introduced by directly reused KV cache can propagate and accumulate across decoding steps, ultimately degrading model accuracy. To address this issue, we argue that it is essential to dynamically eliminate such errors during the decoding phase. 

We find that the attention distributions evolve dynamically throughout the decoding process—an effect we refer to as attention drift (as illustrated in Figure\ref{fig:attention_shift}). Consequently, the column-sum term $\alpha$ used in Algorithm \ref{alg:deviation_aware} also changes over time, making the original importance scores obsolete.

To address this drift, we adapt Algorithm \ref{alg:deviation_aware} for the decoding phase by updating the attention weights to reflect the current attention between the decoding token and the reused tokens. This ensures that attention accuracy is preserved throughout the generation process, enabling more reliable KV cache reuse in multi-stage LLM inference.

As shown in Appendix \ref{appendix:C}, our empirical study demonstrates that this approach effectively stabilizes the model’s output distribution, leading to improved generation quality and reduced perplexity.

\begin{figure*}[htbp]
    \centering
    \begin{subfigure}[t]{0.45\textwidth} 
        \includegraphics[width=0.8\linewidth]{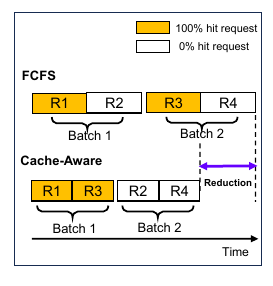}
        \caption{Execution time of different scheduling strategy.}
        \label{fig:exec_batch_hit_a}
    \end{subfigure}
    \hfill
    \begin{subfigure}[t]{0.45\textwidth} 
        \includegraphics[width=0.9\linewidth]{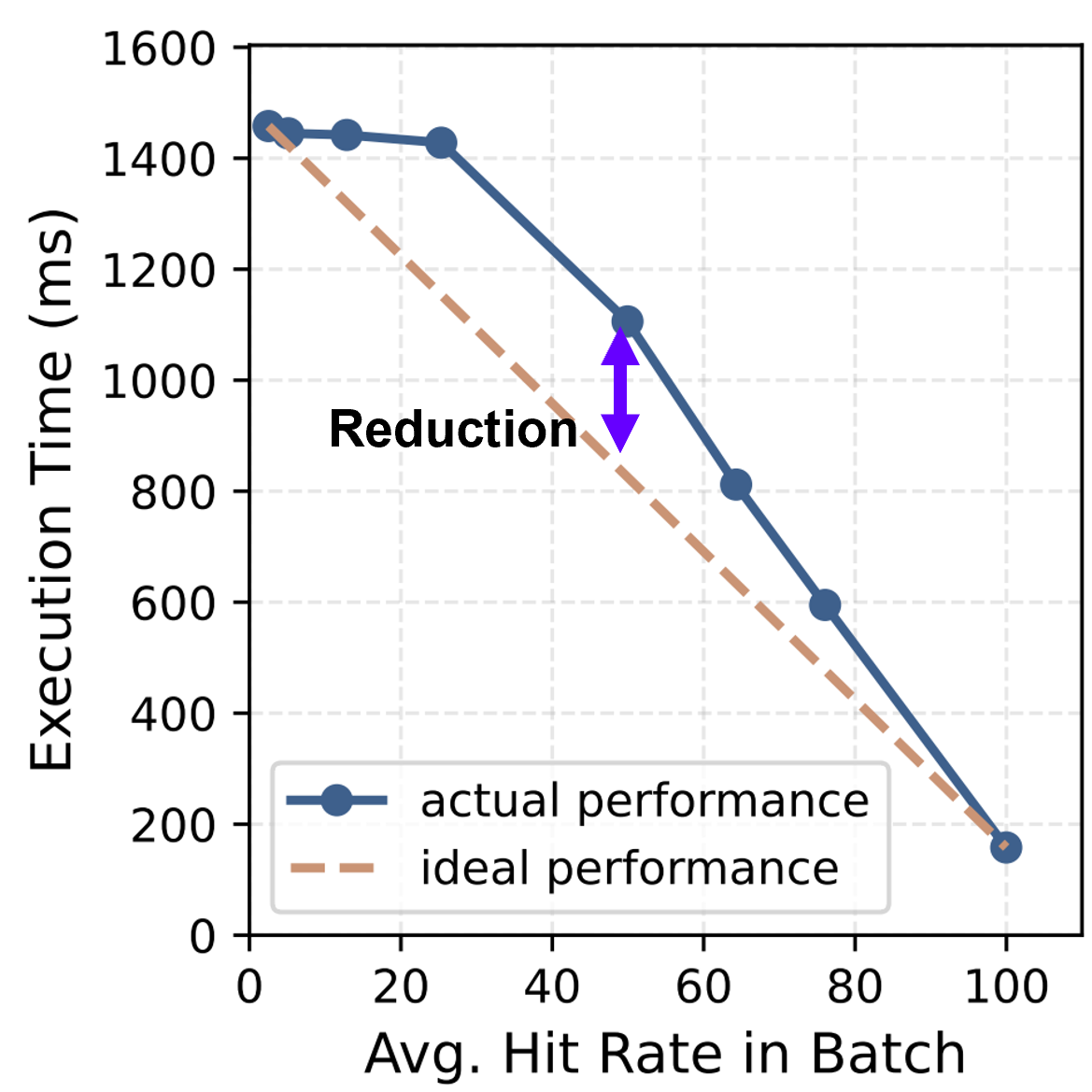}
        \caption{
Batch execution time under varying hit rates.}
        \label{fig:schedule_demo}
    \end{subfigure}
    \caption{Cache-Aware scheduling optimizes Transformer batch processing performance.Figure\ref{fig:exec_batch_hit_a}: Comparison of FCFS and Cache-Aware scheduling: FCFS causes significant hit rate disparities in the same batch (e.g., R1/R3 vs. R2/R4), while Cache-Aware accelerates TTFT by prioritizing high-hit-rate requests and merging similar ones. Figure\ref{fig:schedule_demo}: The gap between actual and ideal performance curves shows Cache-Aware’s TTFT optimization potential (tested on Llama3.1-8b with GSM8K).}

\label{fig:exec_batch_hit}
\end{figure*}

\subsection{Efficient Multi-Request Continuous Batch Scheduling Strategy}
\label{sec:scheduling}

In current LLM multi-tenant inference systems, to fully improve GPU utilization and maximize input throughput, the continuous batching technology is often used to combine multiple prefill requests into one request for calculation. In the multi-tenant system studied in this article, during the calculation process, prefill requests can reduce part of the computational load by hitting the KV Cache of some tokens, thus accelerating the prefill speed.

However, if a conventional scheduling strategy is adopted, prefill requests with high KV hit rates and those with low KV hit rates may be combined into a batch for calculation. As a result, prefill requests with high KV hit rates cannot achieve the expected performance improvement (i.e., cannot effectively reduce TTFT). This is because the longer calculation time of requests with low KV hit rates masks the acceleration benefits brought by requests with high hit rates.

To avoid the batching of high KV hit rates and those with low KV hit rates, we analyze the Time-To-First-Token (TTFT) latency of a batch under varying cache hit rates. The TTFT latency exhibits a concave relationship with respect to the average hit rate, as shown in Figure \ref{fig:exec_batch_hit_a}. This insight motivates the design of our cache-aware scheduling algorithm.
A formal analysis of the algorithm’s theoretical properties is provided in the Appendix \ref{appendix:D}.

\begin{algorithm}[H]
\caption{Request Scheduling Algorithm}
\label{alg:scheduling}
\SetAlgoLined
\KwIn{Incoming request $r_i$, request queue $Q$}
\KwOut{Response for $r_i$}

Compute the cache hit rate $h_i$ for $r_i$ from the KV Cache pool.\

$Q$.insert($r_i$)  \\
$Q$.sort(descending $h_i$) \\
Send top $N$ requests from $Q$ to the LLM, where $N$ is the batch size.\

\Return Response for $r_i$
\end{algorithm}

%% file: section/eval.tex
\section{Experiments}

\subsection{Implementation}
\label{section:5.1}
We implement KVShare based on VLLM 0.4.1 \citep{kwon2023efficient} with about 2K lines of code in Python. We incorporate the schedule method and the DHD algorithm. We port CacheBlend from their public repository. For more details, please refer to Appendix \ref{appendix:A}.

\subsection{Settings}
\label{section:5.2}

 \noindent\textbf{Models and Hardware Settings:} We have evaluated the performance of KVShare in open-source large language models (LLMs) with different model architectures and parameter counts, such as Qwen2.5-7B \citep{qwen2.5}, Llama3.1-8B \citep{llama3modelcard} and Yi1.5-9B \citep{young2024yi}. During the experimental process, we uniformly used one NVIDIA-L40S GPU as the hardware support for the operation of large language models (LLMs). 

\noindent\textbf{Baselines:} We consider several baselines for comparison. The Full recompute baseline completely recomputes all tokens, while the Naive baseline completely reuses the KVCache of tokens. Additionally, we include CacheBlend \citep{yao2024cacheblend} and EPIC \citep{hu2024epic} as SOTA methods.

\noindent\textbf{Datasets:} To comprehensively evaluate our system KVShare, we conduct tests against all the baseline methods across diverse types of tasks. These tasks primarily encompass document summarization, where we use the SAMSum \citep{gliwa-etal-2019-samsum}; mathematical reasoning, leveraging the GSM8K \citep{cobbe2021gsm8k}; DROP (Discrete Reasoning Over Paragraphs) \citep{Dua2019DROP}. 

\noindent\textbf{Evaluation Metrics:} We adopt the following standard metrics to measure the generation quality: \textbf{Accuracy} \citep{hastie2009elements} is used to evaluate GSM8K dataset. It represents the proportion of problems correctly solved by the model. \textbf{EM Score} \citep{Dua2019DROP} is used to evaluate DROP dataset. It measures the proportion of character-level exact matches between the answers generated by the model and the annotated ground truth. \textbf{Rouge-L} \citep{lin2004rouge} is used to evaluate SAMSum dataset. It measures the similarity between the model’s output and the ground-truth summaries based on the longest common sequence. 

\subsection{Overall Improvement}


To rigorously validate the DHD metric and investigate decode-stage token recomputation impacts on LLM accuracy under KV reuse, two controlled experiments compare our method with baselines. The first focuses on the prefill stage, evaluating how DHD-guided token recomputation in KVShare restores accuracy versus baselines. The second examines decode-stage recomputation's incremental accuracy gains over prefill-only recomputation and compares methods' accuracy performance, 
adding 3 additional token recomputations per decode step.

\begin{table}[]
\caption{We tested KVShare and baselines on 3 datasets and 3 models with 4 recomputation ratios, covering prefill-only and prefill+decode recomputation scenarios. The DHD metric enabled our method to nearly match FR (Full Recomputation) in prefill stages, while decode-stage recomputation boosted all methods' accuracy—our approach remained closest to FR via DHD effectiveness.}
\vspace{1em}
\label{tab:performance}

\resizebox{\linewidth}{!}{
\begin{tabular}{c|ccc|ccc|ccc}
\toprule
                & \multicolumn{3}{c}{GSM8K (Accuracy)} & \multicolumn{3}{c}{DROP (EM Score)} & \multicolumn{3}{c}{SAMSum (RougeL)} 
               \\
                & \makecell{Qwen2.5 \\ 7B}  & \makecell{Llama3.1 \\ 8B} & \makecell{Yi1.5\\ 9B} & \makecell{Qwen2.5 \\ 7B} &\makecell{Llama3.1\\8B} & \makecell{Yi1.5\\ 9B} & \makecell{Qwen2.5 \\ 7B} & \makecell{Llama3.1\\ 8B} & \makecell{Yi1.5\\ 9B} \\
            \midrule
Full Recompute  & 82.81       & 54.49       & 69.92    & 60.75      & 69.65       & 68.43    & 20.17      & 18.88       & 18.24    \\
Naive           & 50.00       & 37.79       & 20.70    & 31.25      & 42.96       & 40.78    & 8.39       & 5.05        & 9.28     \\
\cmidrule(lr){1-10}
\multicolumn{10}{c}{Recompute in Prefill Stage}                                                                                    \\
\cmidrule(lr){1-10}
CacheBlend@0.10 & 42.00       & 37.30       & 24.80    & 36.72      & 45.31       & 42.75    & 13.92      & 12.10       & 13.98    \\
CacheBlend@0.20 & 51.10       & 39.64       & 31.25    & 38.28      & 50.00       & 53.90    & 13.92      & 13.05       & 14.19    \\
CacheBlend@0.30 & 54.68       & 42.57       & 32.26    & 40.63      & 57.81       & 53.90    & 15.35      & 13.21       & 14.34    \\
CacheBlend@0.40 & 60.93       & 44.33       & 35.35    & 39.06      & 56.25       & \cellcolor{green!20}{59.37}    & 15.80      & 14.70       & 14.32    \\
EPIC@0.10       & 47.26       & 33.98       & 25.58    & 45.31      & 33.98       & 46.09    & 14.29      & 15.20       & 14.06    \\
EPIC@0.20       & 47.46       & 33.98       & 24.21    & 46.09      & 33.98       & 45.31    & 14.54      & 15.22       & 13.70    \\
EPIC@0.30       & 48.04       & 32.62       & 21.48    & 45.31      & 32.61       & 45.31    & 14.70      & \cellcolor{green!20}{15.67}       & 13.93    \\
EPIC@0.40       & 50.58       & 33.39       & 25.00    & 47.03      & 33.98       & 50.00    & 14.78      & 15.66       & 13.50    \\
KVShare@0.10    & 50.00       & 33.59       & \cellcolor{green!20}{51.56}    & 41.40      & 52.34       & 53.90    & 15.75      & 12.02       & 14.85    \\
KVShare@0.20    & 52.34       & 35.74       & 45.72    & 45.31      & 56.25       & 47.65    & 16.13      & 14.62      & 15.18    \\
KVShare@0.30    & 55.47       & 40.23       & 49.84    & 43.75      & 58.59       & 50.78    & 16.79      & 15.01       & 16.01    \\
KVShare@0.40    & \cellcolor{green!20}{63.28}      & \cellcolor{green!20}{44.53}       & 44.34    & \cellcolor{green!20}{48.43}      & \cellcolor{green!20}{63.28}       & 56.25    & \cellcolor{green!20}{17.21}      & 15.21       & \cellcolor{green!20}{16.03}    \\
\cmidrule(lr){1-10}
\multicolumn{10}{c}{Recompute in Prefill \& Decode Stage}                                                                          \\
\cmidrule(lr){1-10}
CacheBlend@0.10 & 71.86       & 50.78       & 42.77    & 41.40      & 46.87       & 50.00    & 14.74      & 12.01       & 13.80    \\
CacheBlend@0.20 & 71.88       & 51.17       & 42.96    & 42.96      & 54.68       & 53.90    & 14.79      & 13.23       & 14.67    \\
CacheBlend@0.30 & 72.65       & 46.67       & 42.38    & 47.65      & 57.81       & 57.03    & 15.92      & 13.36       & 14.55    \\
CacheBlend@0.40 & 73.22       & 50.19       & 43.55    & 46.09      & 57.81       & 60.15    & 16.20      & 14.90       & 14.47    \\
EPIC@0.10       & 73.25       & 50.78       & 45.89    & 54.68      & 50.78       & 46.09    & 15.16      & 15.31       & 13.82    \\
EPIC@0.20       & 71.39       & 47.07       & 41.40    & 45.31      & 47.07       & 45.31    & 15.64      & 15.34       & 13.99    \\
EPIC@0.30       & 70.21       & 47.07       & 36.91    & 53.90      & 47.07       & 45.31    & 15.62      & 15.55       & 13.77    \\
EPIC@0.40       & 69.53       & 46.09       & 36.91    & 55.48      & 46.09       & 50.00    & 15.71      & 15.77       & 13.89    \\
KVShare@0.10    & \cellcolor{green!40}{75.78}       & 55.86       & 51.56    & 46.87      & 54.68       & 58.59    & 17.18      & 14.05       & 14.68    \\
KVShare@0.20    & 71.88       & 53.52       & 58.39    & 53.90      & 62.50       & 53.90    & 17.89      & 15.78       & 15.14    \\
KVShare@0.30    & 71.48       & \cellcolor{green!40}{58.20}       & 58.78    & 54.92      & 66.40       & 63.28    & 18.10      & 16.35       & 15.89    \\
KVShare@0.40    & 72.27       & 57.04       & \cellcolor{green!40}{62.30}    & \cellcolor{green!40}{56.15}      & \cellcolor{green!40}{68.75}       & \cellcolor{green!40}{65.63}    & \cellcolor{green!40}{19.05}      & \cellcolor{green!40}{17.17}       & \cellcolor{green!40}{16.34}   \\
\bottomrule
\end{tabular}}
\end{table}

In the first set experiments, we found that DHD algorithm demonstrates significant advantages over SOTA methods such as Epic and CacheBlend. As shown in Table \ref{tab:performance}, only when performing recalculation in prefill, KVShare's performance score showed an average increase of 9.48\% compared to Cacheblend and an average increase of 29.98\% compared to Epic. This improvement stems from KVShare's proposed DHD algorithm, which prioritizes fine-grained processing of high-value tokens (e.g., query keywords, logical connectives) during the prefill stage. This approach maintains model generation accuracy while reducing invalid computations.

In the second set of experiments, KVShare's advantages grow with decode-stage recomputation. Table \ref{tab:performance} shows adding decode-stage recomputation yields 16.76\% average performance improvement over prefill-only recomputation across scenarios, with KVShare outperforming SOTA methods by 20.38\% on average. The improvement originates from its two-stage mechanism: Prefill stage mitigates high-impact attention-error tokens via recomputation; Decode stage reduces KV-error token influence through incremental stepwise recomputation. Their collaboration enhances cross-request KV cache reuse accuracy and balances efficiency-quality tradeoffs.


\subsection{Effectiveness Analysis}

To verify the performance of the KVShare system in real multi-tenant scenarios, we extracted real human instructions from datasets such as ShareGPT \citep{ShareGPT-Chinese-English-90k} and regulated the sending rate to test the improvement of token reuse on TTFT and input throughput in multi-tenant environments. Based on the real VLLM system, we compared the performance of KVShare with CacheBlend and EPIC systems under multi-tenant conditions.


\begin{figure*}[htbp]
    \vspace{-0.5em} 
    \centering
    \begin{subfigure}[t]{0.4\textwidth} 
        \centering
        \includegraphics[width=\linewidth, height=5cm, keepaspectratio]{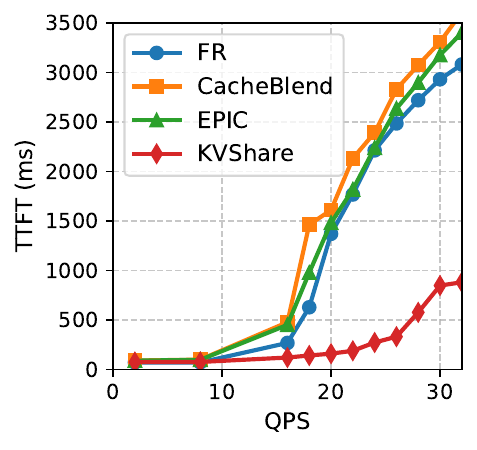}
        \caption{TTFT}
        \label{fig:ttft_comparison}
    \end{subfigure}
    \hspace{1em} 
    \begin{subfigure}[t]{0.4\textwidth} 
        \centering
        \includegraphics[width=\linewidth, height=5cm, keepaspectratio]{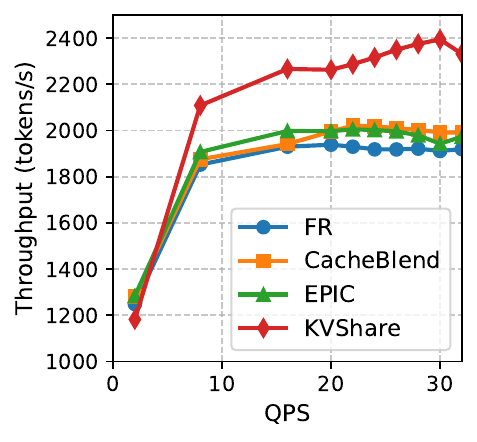}
        \caption{Throughput}
        \label{fig:throughput_comparison}
    \end{subfigure}
    \caption{Efficiency analysis for KVshare. (a) and (b) show TTFT and Throughput under 20\% recomputation ratio.} 
    \vspace{-0.5em} 
\end{figure*}

In terms of the TTFT metric, as shown in Figure \ref{fig:ttft_comparison}, KVShare exhibits substantial low-latency advantages and load robustness when compared with the full recomputation method (FR) and SOTA approaches EPIC and CacheBlend. 
As the query rate increases, the TTFT of FR, Epic, and CacheBlend grows exponentially owing to a dramatic decline in cache hit rates. In contrast, even with a high concurrency level of 30 QPS, KVShare stabilizes TTFT below 881 ms.
This improvement stems from KVShare's cache-aware scheduling strategy: requests are batched by KV hit rates, separating high and low hit-rate requests to avoid the "tail latency effect" where full recomputation of low hit-rate requests delays high hit-rate ones, thus achieving the optimal balance between TTFT and resource utilization.

In the throughput comparison experiments, as shown in the Figure \ref{fig:throughput_comparison}, when the QPS is low, the throughput of each method is similar. When the QPS exceeds 8, KVShare's throughput significantly outperforms other methods, with an average 1.2× improvement over FR. As the QPS increases, the throughput of other methods declines to varying degrees, while KVShare maintains stable performance even at QPS=30. This demonstrates the effectiveness of KVShare in multi-tenant concurrent scenarios, primarily due to its dynamic optimization of cross-request KV cache reuse efficiency through a fine-grained cache sharing mechanism.

\begin{figure*}[htbp]
        \includegraphics[width=\linewidth]{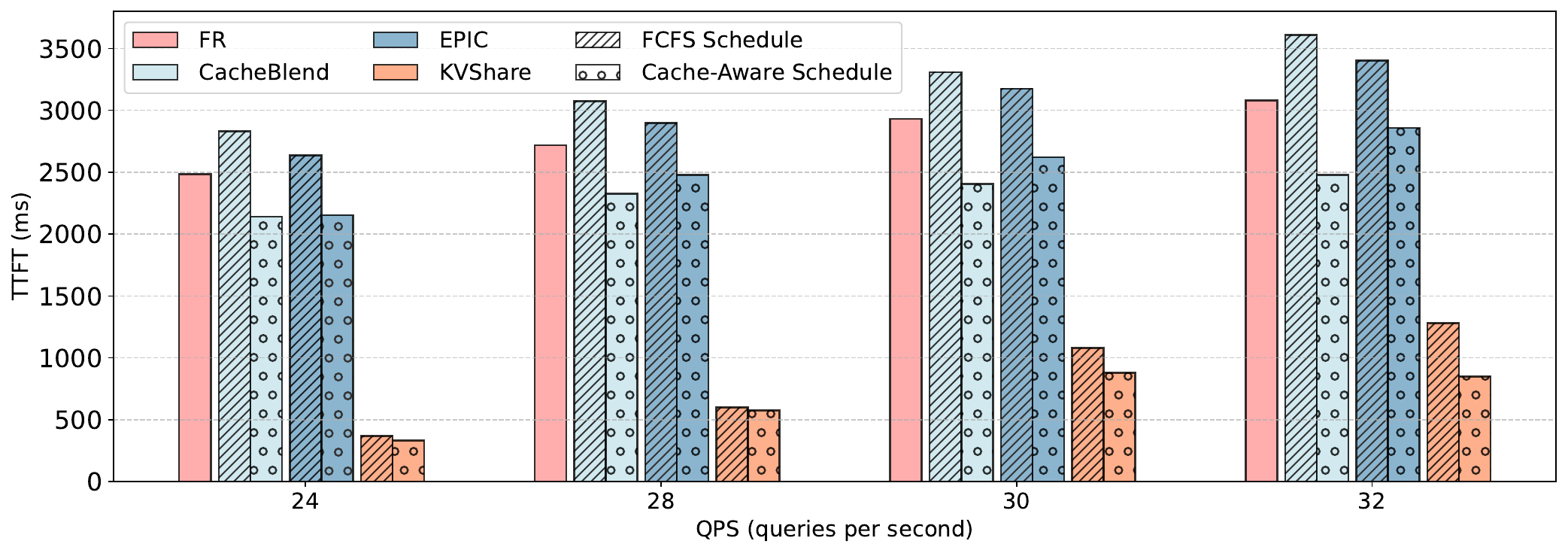}
        \caption{TTFT of KVShare, Epic, and CacheBlend without cache-aware scheduling strategy against the same methods with the strategy enabled.}
        \label{fig:ttft_vs_qps_methods}
\end{figure*}

We verified the effectiveness of the cache-aware scheduling strategy in reducing the TTFT of multi-tenant requests through ablation experiments. The experiment compared the KVShare, Epic, and CacheBlend methods without this strategy enabled with the same methods with the strategy enabled. As shown in Figure \ref{fig:ttft_vs_qps_methods}, in different QPS scenarios, the methods with the strategy enabled significantly reduce the TTFT, with a maximum reduction of 33.8\%. In the multi-tenant KV cache reuse scenario, the strategy processes requests with similar token hit rates in groups, avoiding the performance degradation caused by the mixed scheduling of requests with high and low hit rates, thus effectively reducing the TTFT. 

%% file: section/conclusion.tex
\section{Conclusion}

This paper introduces KVShare, an efficient and effective multi-tenant KV cache sharing framework. By using a cache-aware scheduler and DHD algorithm, it reduces TTFT by up to 9.39×, boosts throughput by 20\%, and outperforms SOTA methods. KVShare shows robust high-concurrency performance and paves the way for efficient and sustainable LLM services.

%% file: section/appendix.tex
\newpage
\section*{Limitations}
\label{Limitations}

This work for the first time proposes the token recomputation mechanism during both the prefilling and decoding phase, which corrects potential deviations from KV reused tokens in prefill by introducing the Dual-stage High Deviation (DHD) algorithm. However, the TPOT (Time Per Output Token) of the LLM system increased because adding N additional DHD token computations per decoding step under KVShare’s approach doubles TPOT across QPS scenarios due to increased DECODE request load, with each additional token introducing approximately a 1x performance penalty, as shown in Figure \ref{fig:kvshare_overhead}. Future work will focus on further mitigating the impact on TPOT by reducing this load via efficient scheduling and selective token recomputation.


\section*{Broader Impacts}
\label{Broader Impacts}

This work proposes an adaptive caching and computation reuse framework breaks through the cache efficiency bottleneck, enhancing multi-tenant resource utilization, reducing energy consumption costs in cloud or edge computing, and promoting the inclusive deployment of AI in civilian sectors such as healthcare and education. Its low-latency characteristics facilitate real-time interactive applications, while optimized resource utilization aligns with green computing goals. The research mitigates technological monopoly risks through open-sourcing algorithms and standardized documentation. Its methodology is expected to extend to distributed computing and other fields, fostering an inclusive and sustainable intelligent computing ecosystem.
\appendix

\section{System implementation}
\label{appendix:A}

\begin{figure}[htbp]
    \centering  
    \begin{minipage}[b]{0.45\textwidth}
        \centering
        \includegraphics[width=\linewidth,height=6cm]{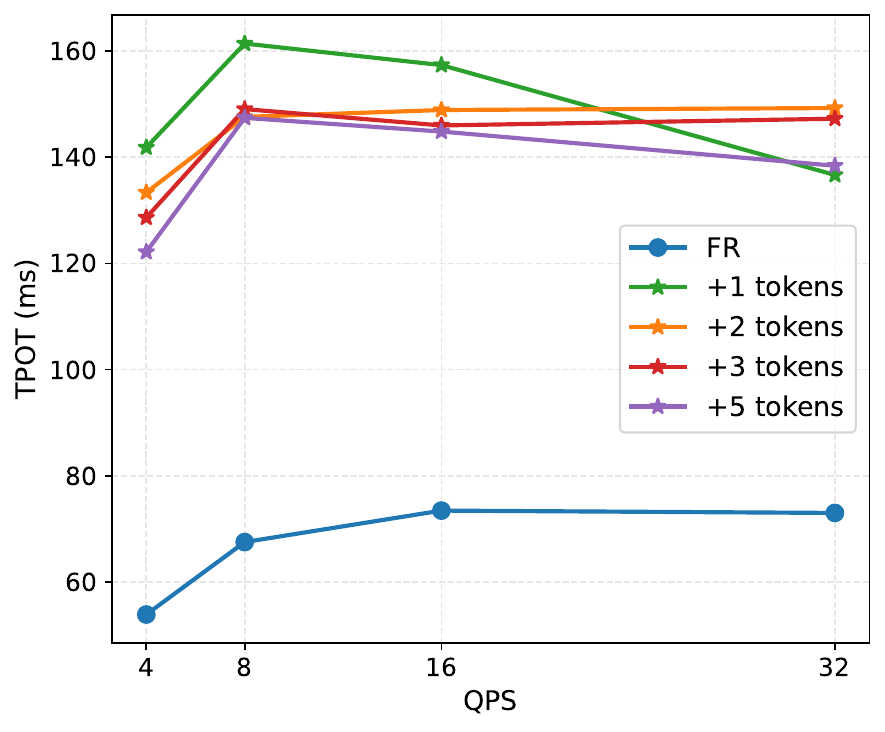}
        \caption{Comparison of TPOT between FR Mode and KVShare with Additional DHD Tokens per Decoding Step.}
        \label{fig:kvshare_overhead}
    \end{minipage}
    \hfill 
    \begin{minipage}[b]{0.45\textwidth}
        \centering
        \includegraphics[width=\linewidth,height=6cm]{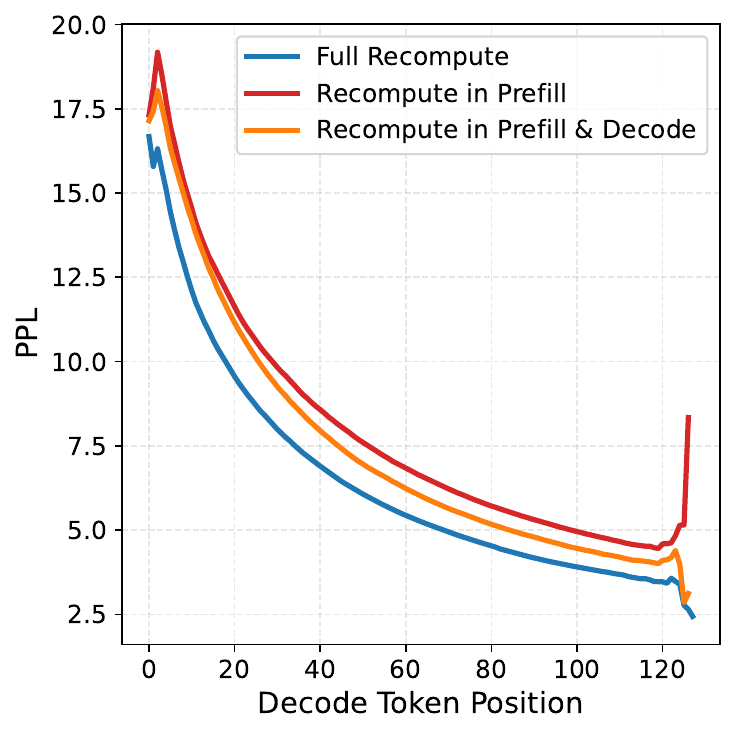}
        \caption{Impact of Recomputing Strategies on Perplexity During Decoding (Evaluated on GSM8K).}
        \label{fig:ppl_analysis_plot}
    \end{minipage}
    \label{fig:appendix1}
\end{figure}

Our implementation in the open-source vllm codebase fully integrates the KVShare system, as shown in Figure\ref{fig:kvshare_system}, which consists of four main components: KV Retriever, Cache-Aware Scheduler, DHD Selector, and KV Writer. Upon receiving a new user request, the KV Retriever employs a rolling hash algorithm to identify and match reusable token segments from previous requests, loading the corresponding KV caches from disk into the KV cache pool. Each request is then assigned a token hit rate after processing by the KV Retriever.

The Cache-Aware Scheduler prioritizes requests in the scheduling queue based on their token hit rates in descending order, with higher hit rates receiving precedence. Requests with similar hit rates are batched together to minimize the average Time-To-First-Token (TTFT) for prefill operations.

During LLM execution, the DHD Selector dynamically identifies Dynamic Hash Decision (DHD) tokens based on attention scores, allowing the model to compute only these critical tokens while reusing KV caches from others. This strategy optimizes computation efficiency while maintaining minimal prefill accuracy loss.

In the decoding phase, the DHD Selector continues to select DHD tokens (specifically those reused during prefill but not computed) to further reduce accuracy degradation caused by KV reuse through targeted recomputation.

\section{Empirical Analysis of Recomputing Strategies in the Decoding Phase}
\label{appendix:C}

As shown in the Figure \ref{fig:ppl_analysis_plot}, the Recompute in Prefill \& Decode strategy consistently achieves lower perplexity compared to recomputing in the prefill stage only, and closely approximates the performance of full recomputation.
This demonstrates that selective recomputation during the decoding phase is crucial for mitigating cumulative attention bias, effectively improving the model’s generation quality.

\section{How to selecte Token indices}
\label{appendix:B}
    \subsection{The positions of DHD tokens remarkably overlap across successive layers}
    \label{appendix:D.1}

\begin{figure*}[htbp]
    \centering  
    \begin{minipage}[t]{0.45\textwidth}
        \centering
        \includegraphics[width=\linewidth,height=4.8cm]{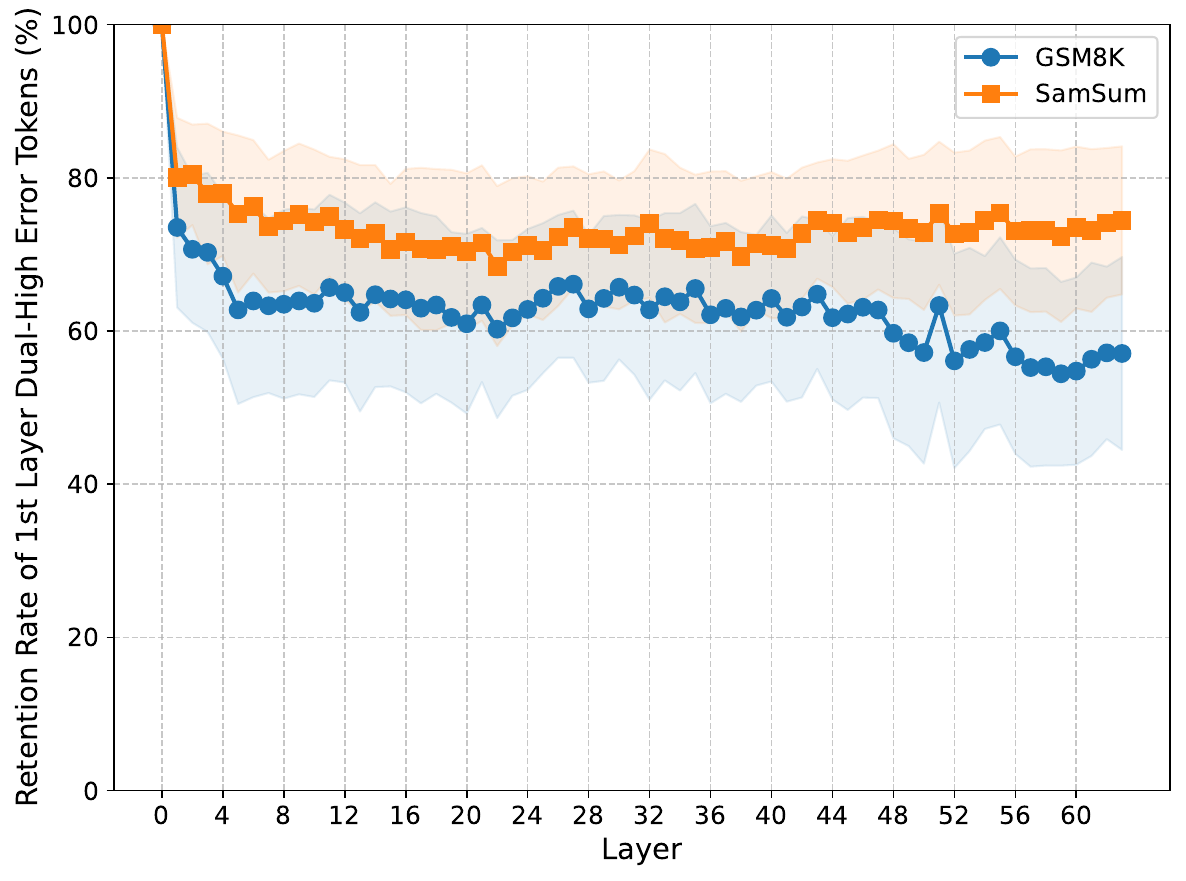}
        \caption{The retention rate of dual high error tokens in the first layer of the Qwen2.5-32B model in subsequent different layers.}
        \label{fig:visual_case_dhd token_token2}
    \end{minipage}
    \hfill 
    \begin{minipage}[t]{0.45\textwidth}
        \centering
        \includegraphics[width=\linewidth,height=5.4cm]{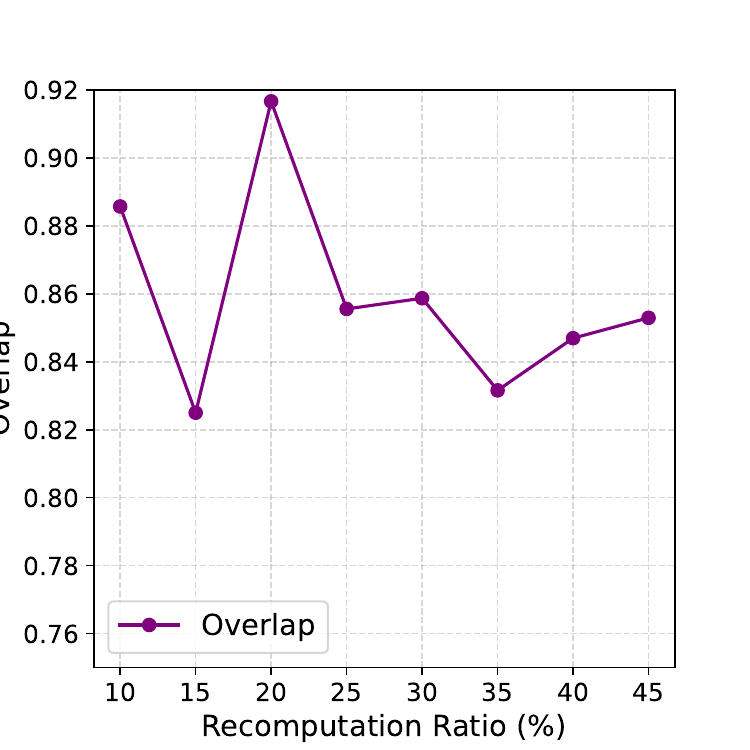}
        \caption{Token indices selected based on $K$ and $V$ impacts consistently exhibit 80\% to 90\% overlap.}
        \label{fig:overlap between K impact and V impact}
    \end{minipage}
    \label{fig:appendix2}
\end{figure*}

In this section, We analyzed the propagation effects of DHD Tokens across different layers of large language models (LLMs). We quantify this phenomenon in Figure \ref{fig:visual_case_dhd token_token2},, showing that 60\%-80\% of DHD Tokens computed in the first layer persist across the subsequent N layers across various datasets. This observation motivates a novel optimization strategy: by precomputing DHD Token positions in the early layers of an LLM, we can selectively recompute only these critical tokens in subsequent layers to reduce \(\Delta A\), thereby maintaining model accuracy while enhancing computational efficiency during the prefill phase.

\subsection{The token indices selected based on K and V impacts consistently exhibit overlap}
\label{appendix:D.2}

As shown in Figure \ref{fig:overlap between K impact and V impact}, we first pre-collected the KV Cache of prompt on the GSM8K dataset and analyzed the impact of KV errors and real attention output errors using the formula \eqref{eq:attention_differentiation}. We found that the token indices selected based on K and V impacts consistently exhibit 80\% to 90\% overlap. 

\section{Proof of the Schedule method}
\label{appendix:D}

\paragraph{Local Optimality Condition} 
Let $\mathcal{B} = \{B_1, \dots, B_k\}$ be a batch configuration sorted in descending order of KV cache hit rates. If for any pair of batches $B_i$, $B_j$ ($i < j$) and arbitrary request exchange $r_i \leftrightarrow r_j$, the total Time-To-First-Token (TFT) latency satisfies:

\begin{equation}
\begin{aligned}
\mathcal{L}(\mathcal{B}) \leq \mathcal{L}(\mathcal{B}')
\end{aligned}
\end{equation}

where $\mathcal{B}'$ denotes the modified configuration after exchange, then the sorting strategy constitutes a local optimum.

\paragraph{Global Optimality} 
Given the concavity of TFT latency $\mathcal{L}$ with respect to request hit rates $\{h_r\}_{r\in\mathcal{R}}$, the local optimality condition in Proposition 1 guarantees global optimality of the sorted batching strategy.

\subsubsection*{Formal Proof Framework}

\begin{enumerate}
    \item \textbf{Notation Setup}
    Let $ f: \mathbb{R}^+ \to \mathbb{R} $ denote the concave latency function for an individual batch. The batch hit rate is defined by:
    \begin{equation}
\begin{aligned}
    H(B) := \frac{1}{|B|}\sum_{r \in B} h_r
\end{aligned}
\end{equation}
    \item \textbf{Key Inequality}
    For any two batches $B_1$, $B_2$ with $H(B_1) > H(B_2)$ and perturbation $\delta > 0$:
    \begin{equation}
\begin{aligned}
    f(H(B_1)) + f(H(B_2)) \leq f(H(B_1) - \delta) + f(H(B_2) + \delta)
\end{aligned}
\end{equation}
    \textit{Proof}: Direct application of concave function properties .
    
    \item \textbf{Latency Monotonicity}
    The total latency $\mathcal{L}(\mathcal{B}) = \sum_{i=1}^k f(H(B_i))$ is minimized when batches are sorted by $H(B_i)$.
\end{enumerate}



\section{Adaptive-length token reuse algorithm }
\label{appendix:E}

\begin{algorithm}[H]
\caption{Efficient Token Sequence Matching with Rolling Hash}
\label{alg:token_matching}
\SetAlgoLined
\KwIn{\textit{target\_tokens} (Target token sequence), \textit{candidate\_tokens} (Candidate token sequence), \textit{window\_size} (Sliding window size)}
\KwOut{\textit{target\_matches} (Matched positions in target), \textit{candidate\_matches} (Matched positions in candidate)}

\tcc{Step 1: Initialize result sets}
$target\_matches \leftarrow []$, $candidate\_matches \leftarrow []$ \\
$matched\_indices \leftarrow \emptyset$ \\

\tcc{Step 2: Build target token hash index}
$target\_hash\_index \leftarrow \emptyset$ \\
\For{$i = 0$ \textbf{to} $\text{len}(target\_tokens) - window\_size$}{
    $h \leftarrow \text{ComputeRollingHash}(target\_tokens, i, window\_size)$ \\
    $target\_hash\_index[h] \leftarrow target\_hash\_index.get(h, []) + [i]$ 
}

\tcc{Step 3: Match candidate tokens using hash index}
\For{$j = 0$ \textbf{to} $\text{len}(candidate\_tokens) - window\_size$}{
    $h \leftarrow \text{ComputeRollingHash}(candidate\_tokens, j, window\_size)$ \\
    \For{$i \in target\_hash\_index.get(h, [])$}{
        $k \leftarrow 0$ \\
        \While{$i+k < \text{len}(target\_tokens)$ \textbf{and} $j+k < \text{len}(candidate\_tokens)$ \textbf{and} $target\_tokens[i+k] = candidate\_tokens[j+k]$}{
            \If{$(i+k) \notin matched\_indices$}{
                $target\_matches.\text{append}(i+k)$ \\
                $candidate\_matches.\text{append}(j+k)$ \\
                $matched\_indices.\text{add}(i+k)$
            }
            $k \leftarrow k + 1$
        }
    }
}

\tcc{Step 4: Return matched positions}
\Return $(target\_matches, candidate\_matches)$
\end{algorithm}

\paragraph{Polynomial Hash Function} 
The rolling hash used in this algorithm is a polynomial hash function defined as:
\begin{equation}
H = (t_0 \cdot b^{w-1} + t_1 \cdot b^{w-2} + \cdots + t_{w-1} \cdot b^0) \bmod m
\end{equation}
where $t_i$ are the tokens in the current window, $b$ is the base, $w$ is the window size, and $m$ is the modulus. This function efficiently computes hash values for fixed-length token windows, allowing for fast matching while minimizing collisions.

%% file: section/checklist.tex
\newpage
\section*{NeurIPS Paper Checklist}

\begin{enumerate}

\item {\bf Claims}
    \item[] Question: Do the main claims made in the abstract and introduction accurately reflect the paper's contributions and scope?
    \item[] Answer: \answerYes{} 
    \item[] Justification: We provide detailed experimental results to support our claims.

\item {\bf Limitations}
    \item[] Question: Does the paper discuss the limitations of the work performed by the authors?
    \item[] Answer: \answerYes{} 
    \item[] Justification: We discuss the limitations of RankRAG in Appendix \ref{Limitations}.

\item {\bf Theory assumptions and proofs}
    \item[] Question: For each theoretical result, does the paper provide the full set of assumptions and a complete (and correct) proof?
    \item[] Answer: \answerYes{} 
    \item[] Justification: We provide a complete proof in Section \ref{section:4.1} and Appendix \ref{appendix:D}.

    \item {\bf Experimental result reproducibility}
    \item[] Question: Does the paper fully disclose all the information needed to reproduce the main experimental results of the paper to the extent that it affects the main claims and/or conclusions of the paper (regardless of whether the code and data are provided or not)?
    \item[] Answer: \answerYes{} 
    \item[] Justification: We provide implementation details in Section \ref{section:5.1} and Appendix \ref{appendix:A}. Besides, we provide the source code used in this paper.

\item {\bf Open access to data and code}
    \item[] Question: Does the paper provide open access to the data and code, with sufficient instructions to faithfully reproduce the main experimental results, as described in supplemental material?
    \item[] Answer: \answerYes{} 
    \item[] Justification: Our source code is public available data with open access.

\item {\bf Experimental setting/details}
    \item[] Question: Does the paper specify all the training and test details (e.g., data splits, hyperparameters, how they were chosen, type of optimizer, etc.) necessary to understand the results?
    \item[] Answer: \answerYes{} 
    \item[] Justification: Our Experimental settings can be found in Section \ref{section:5.2}

\item {\bf Experiment statistical significance}
    \item[] Question: Does the paper report error bars suitably and correctly defined or other appropriate information about the statistical significance of the experiments?
    \item[] Answer: \answerNo{} 
    \item[] Justification: The paper provides evaluation results and error analysis.


\item {\bf Experiments compute resources}
    \item[] Question: For each experiment, does the paper provide sufficient information on the computer resources (type of compute workers, memory, time of execution) needed to reproduce the experiments?
    \item[] Answer: \answerYes{} 
    \item[] Justification: We provide the compute resources information in Section \ref{section:5.2}.
    
\item {\bf Code of ethics}
    \item[] Question: Does the research conducted in the paper conform, in every respect, with the NeurIPS Code of Ethics \url{https://neurips.cc/public/EthicsGuidelines}?
    \item[] Answer: \answerYes{} 
    \item[] Justification: We conform with the NeurIPS Code of Ethics.

\item {\bf Broader impacts}
    \item[] Question: Does the paper discuss both potential positive societal impacts and negative societal impacts of the work performed?
    \item[] Answer: \answerYes{} 
    \item[] Justification: We discuss both potential societal impacts and negative impacts in Appendix \ref{Broader Impacts}
    
\item {\bf Safeguards}
    \item[] Question: Does the paper describe safeguards that have been put in place for responsible release of data or models that have a high risk for misuse (e.g., pretrained language models, image generators, or scraped datasets)?
    \item[] Answer: \answerNA{} 
    \item[] Justification: This paper poses no such risks.

\item {\bf Licenses for existing assets}
    \item[] Question: Are the creators or original owners of assets (e.g., code, data, models), used in the paper, properly credited and are the license and terms of use explicitly mentioned and properly respected?
    \item[] Answer: \answerYes{} 
    \item[] Justification:  We properly cite the data, and model used in this paper.

\item {\bf New assets}
    \item[] Question: Are new assets introduced in the paper well documented and is the documentation provided alongside the assets?
    \item[] Answer: \answerYes{} 
    \item[] Justification: We provide detailed readme to the open-sourced code.

\item {\bf Crowdsourcing and research with human subjects}
    \item[] Question: For crowdsourcing experiments and research with human subjects, does the paper include the full text of instructions given to participants and screenshots, if applicable, as well as details about compensation (if any)? 
    \item[] Answer: \answerNA{} 
    \item[] Justification: This paper does not involve crowdsourcing nor research with human subjects.

\item {\bf Institutional review board (IRB) approvals or equivalent for research with human subjects}
    \item[] Question: Does the paper describe potential risks incurred by study participants, whether such risks were disclosed to the subjects, and whether Institutional Review Board (IRB) approvals (or an equivalent approval/review based on the requirements of your country or institution) were obtained?
    \item[] Answer: \answerNA{} 
    \item[] Justification: This paper does not involve crowdsourcing nor research with human subjects.

\item {\bf Declaration of LLM usage}
    \item[] Question: Does the paper describe the usage of LLMs if it is an important, original, or non-standard component of the core methods in this research? Note that if the LLM is used only for writing, editing, or formatting purposes and does not impact the core methodology, scientific rigorousness, or originality of the research, declaration is not required.
    \item[] Answer: \answerYes{} 
    \item[] Justification: We describe the usage of LLMs in Appendix \ref{appendix:A}

\end{enumerate}